\documentclass[10pt,twocolumn,letterpaper]{article}

\usepackage[pagenumbers]{cvpr} 

\usepackage{graphicx}
\usepackage{amsmath}
\usepackage{amssymb}
\usepackage{booktabs}
\usepackage{algorithm}
\usepackage{algorithmic}

\usepackage[pagebackref,breaklinks,colorlinks]{hyperref}

\usepackage[capitalize]{cleveref}
\crefname{section}{Sec.}{Secs.}
\Crefname{section}{Section}{Sections}
\Crefname{table}{Table}{Tables}
\crefname{table}{Tab.}{Tabs.}

\def\cvprPaperID{11719} 
\def\confName{CVPR}
\def\confYear{2023}

\usepackage[pagenumbers]{cvpr} 

\usepackage{graphicx}
\usepackage{amsmath}
\usepackage{amssymb}
\usepackage{booktabs}
\usepackage{algorithm}
\usepackage{algorithmic}

\usepackage[pagebackref,breaklinks,colorlinks]{hyperref}

\usepackage[capitalize]{cleveref}
\crefname{section}{Sec.}{Secs.}
\Crefname{section}{Section}{Sections}
\Crefname{table}{Table}{Tables}
\crefname{table}{Tab.}{Tabs.}

\def\cvprPaperID{11719} 
\def\confName{CVPR}
\def\confYear{2023}

\begin{document}

\title{MAViC: Multimodal Active Learning for Video Captioning}



\author{
Gyanendra Das\textsuperscript{\rm 1}\thanks{Work done during internship at ShareChat, India.},
Xavier Thomas\textsuperscript{\rm 1}\footnotemark[1],
Anant Raj\textsuperscript{\rm 2},
Vikram Gupta\textsuperscript{\rm 1}
\\
\textsuperscript{\rm 1}ShareChat, India,
\textsuperscript{\rm 2}INRIA, UIUC\\
\{gyanendradas, xavier.thomas, vikramgupta\}@sharechat.co\\
anant.raj@inria.fr
}
\maketitle

\begin{abstract}
A large number of annotated video-caption pairs are required for training video captioning models, resulting in high annotation costs. Active learning can be instrumental in reducing these annotation requirements. However, active learning for video captioning is challenging because multiple semantically similar captions are valid for a video, resulting in high entropy outputs even for less-informative samples. Moreover, video captioning algorithms are multimodal in nature with a visual encoder and language decoder. Further, the sequential and combinatorial nature of the output makes the problem even more challenging. In this paper, we introduce \textbf{MAViC} which leverages our proposed Multimodal Semantics Aware Sequential Entropy (M-SASE) based acquisition function to address the challenges of active learning approaches for video captioning. Our approach integrates semantic similarity and uncertainty of both visual and language dimensions in the acquisition function. Our detailed experiments empirically demonstrate the efficacy of M-SASE for active learning for video captioning and improve on the baselines by a large margin.


\end{abstract}



\section{Introduction}
\label{sec:intro}
Video content has grown exponentially over the last decade with the rapid adoption of social media platforms, penetration of high end mobile phones and high internet speeds. Availability of large-scale video content and real-world use-cases like self-driving cars, video surveillance, video commerce etc. has attracted lot of interest towards video understanding algorithms. However, training models for these algorithms require large amount of annotated data which is both expensive and time-consuming to collect. To solve the scarcity of \textit{annotated-data}, research community has focused its attention towards various data-efficient algorithms like active learning~\cite{zhan2022comparative, ren2021survey}, semi-supervised learning~\cite{yang2021survey}, zero/few-shot learning~\cite{song2022comprehensive, yang2021survey}, etc. 

In this work, we focus our attention towards active learning for video captioning. Video captioning summarises videos into human-understandable textual sentences which makes these videos more accessible to blind people~\cite{yoon2019video} and also make it easier to search relevant videos. Unlike tasks like video classification and object detection, where the classification taxonomy is closed-set, video captioning offers an open-vocabulary paradigm allowing for deeper video understanding. This makes video captioning a challenging problem because the algorithms should identify the actors present in the video along with the interactions and then describe the overall scene in natural language. Thus, video captioning comprises of a challenging visual understanding and language decoding problem. Unsurprisingly, training a video captioning algorithms require large amount of annotated training data and demands an effective active learning algorithm to reduce the annotation cost and time. While active learning has been studied for various applications like action recognition~\cite{yang2003automatically}, image classification~\cite{collins2008towards}, object detection~\cite{vijayanarasimhan2014large}, human pose estimation~\cite{liu2017active}, natural language processing~\cite{zhang2022survey} etc., active learning for video captioning is still under-explored~\cite{chan2020active}. To fill this gap, we are exploring various techniques for developing effective state-of-the-art active learning techniques for video captioning. 


Active learning for video captioning tasks is extremely challenging because of the following reasons:
\begin{itemize}
    \item \textbf{Sequential output:} Output of video captioning algorithms consist of sequence of words which are generated conditionally over the previous outputs. Thus, commonly used acquisition functions like \textit{entropy}, \textit{margin} need to be adapted to sequential outputs.
    \item \textbf{Multiple correct outputs:} Multiple captions are plausible for a video because captions are not mutually exclusive. For example, \textit{dog running after cat} and \textit{kitten being chased by dog} are two plausible captions for a situation. Acquisition functions based on token-level entropy will generate high entropy for this example even though the example might be less-informative for the model. Thus, active learning algorithms for this task should incorporate semantic similarity in the activation function.
    \item \textbf{Multimodal:} Uncertainty of video captioning algorithms resides in both visual and language modalities. For example, a video consisting of \textit{animated animals} contains lot of visual information, even though the caption might be similar to other videos containing \textit{real animals}. Hence, it is important to incorporate uncertainty from both the modalities.
\end{itemize}

To address the above mentioned challenges, we propose our novel active learning method for video captioning - \textbf{MAViC}. \textbf{MAViC} addresses the sequential nature by extending token level acquisition functions to sequences. To solve the problem of multiple captions, \textbf{MAViC} integrates semantic similarity between candidates in the acquisition function. Thus, semantically similar captions do not dominate the acquisition function. Further, \textbf{MAViC} applies acquisition function on uncertainties computed by both the visual and language part to address the multimodal nature of video captioning task. Combining all these techniques, \textbf{MAViC} is able to improve upon the state-of-the-art by large margin.

\begin{figure*}[t]
  \centering
   \includegraphics[width=\linewidth]{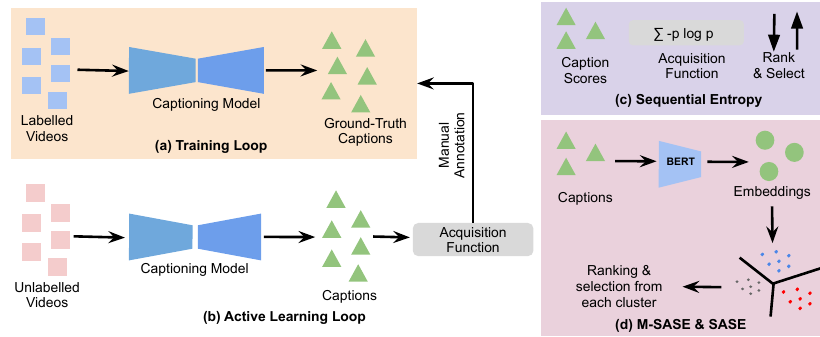}
   \vspace{-2mm}
    \caption{Overall architecture: (a) Supervised training loop (b) Active Learning selects unlabelled datapoints to be sent for manual annotation using acquisition function (c) Baseline Sequential Entropy (SE) approach to rank and query datapoints using sequential entropy (d) Proposed approach to rank and select datapoints using Semantically Aware Sequential Entropy (SASE) and multimodal SASE (M-SASE)}
   \label{fig:msase}
\end{figure*}

\section{Related Work}

\subsection{Video Captioning}
Video captioning~\cite{aafaq2019spatio, chen2019deep, li2021value, liu2018sibnet, zhang2021open, shi2020learning, pei2019memory} has generated lot of interests by the research community and industry. The progress in this field has been accelerated by the contribution of large-scale video captioning datasets like MSR-VTT~\cite{xu2016msr}, MSVD~\cite{chen2011collecting}, MPII-MD~\cite{rohrbach2015dataset}, Charades-STA~\cite{sigurdsson2016hollywood}, ANet-Captions~\cite{krishna2017dense}, LSMDC~\cite{rohrbach2017movie}, VATEX~\cite{wang2019vatex}, YouCook~\cite{ZhXuCoAAAI18} etc. Early attempts at video captioning leveraged Long-Short Term Memory (LSTM) coupled with conditional random fields (CRF)~\cite{DonahueHGRVSD14}, 2D Convolution Neural Networks (CNN) coupled with LSTMs~\cite{VenugopalanXDRMS14} for end-to-end learning, sequence-to-sequence approaches~\cite{Venugopalan_2015_ICCV, SrivastavaMS15} for handling variable length of videos and captions,  3D CNNs for improved spatio-temporal modelling~\cite{yao2015describing}, Fourier transform for improved visual learning~\cite{abs-1902-10322}. However, more recently, methods based on transformer models have shown improvements over previous approaches. Transformer-based approaches pursue different paradigms where some of these freeze the visual encoder and focus on the language decoding step ~\cite{pei2019memory, aafaq2019spatio, pan2020spatio, luo2020univl} while other approaches refine the visual representations also to train the models end-to-end~\cite{chen2019deep, liu2018sibnet, li2021value, zhang2021open,lin2022swinbert}. In this work, we leverage SwinBERT~\cite{lin2022swinbert} for our experiments as it has shown state-of-the-art performance in video captioning. SwinBERT is trained in an end-to-end fashion and uses transformer models for both visual encoding and language decoding. 

\subsection{Active Learning}
Due to high labelling cost of annotated data, active learning has garnered lot of interest from the research community across various applications in computer vision like action recognition~\cite{yang2003automatically}, image classification~\cite{collins2008towards}, object detection~\cite{vijayanarasimhan2014large}, human pose estimation~\cite{liu2017active}, natural language processing~\cite{zhang2022survey} etc. 
There has been a vast amount of work done in the field of active learning and we only provide a brief
overview here. In \cite{settles2012active}, authors provide a detailed survey of studies in the area of active learning before the advent of deep neural networks. Active learning gained even more popularity with deep neural networks and detailed surveys on deep active learning can be found in~\cite{zhan2022comparative, ren2021survey}.  Uncertainty sampling based active learning algorithms are amongst the most popular active learning algorithms. It was introduced by \cite{lewis1994sequential} who empirically justified that uncertainty sampling can improve the performance of text classification substantially. Since then, it has been widely used for performing active learning for various data domain \cite{yang2015multi,zhu2008active,lughofer2017online,yang2016active,wang2017uncertainty,raj2022convergence}. 

Another paradigm of active learning is disagreement based active learning where multiple empirical risk minimizers are maintained and a label is queried if two minimizers disagree on the predicted label for the input sample. A survey of disagreement based active learning is provided in \cite{hanneke2014theory}. These methods has been modified with domain specific heuristics and also applied to visual \cite{wu2022deep,gal2017deep} and language data \cite{arora2007active,schroder2020survey,dor2020active}.  Active learning has a rich literature for image classification \cite{holub2008entropy,joshi2009multi,li2013adaptive,liu2021influence}, object detection \cite{vijayanarasimhan2014large,roy2018deep} and other computer vision tasks as well \cite{siddiqui2020viewal,liu2017active,feng2019deep}. However, active learning for video captioning is relatively an unexplored field. In \cite{chan2020active}, the authors utilize a query by committee (QBC) with cluster-regularization to perform active learning for video captioning task. We discuss more details and differences of our work with ~\cite{chan2020active} in section~\ref{sec:seq_entr}. 

Deep active learning using adversarial perturbation proposes another set of method for active learning. In this setting, the idea is to perturb the model and acquire examples which change their labels on this perturbation. 
Approach proposed by \cite{ducoffe2018adversarial} is closely related to our work. We provide more details in section~\ref{sec:msase-fp}. In natural language processing (NLP) domain, there has been a vast amount of work on active learning for text classification \cite{schroder2020survey}, named-entity recognition \cite{shen2017deep}, semantic role labelling \cite{myers2021tuning}, word sense disambiguation \cite{zhu2007active} etc. In \cite{zhang2022survey}, authors provide an excellent survey of modern active learning methods for NLP tasks. Active learning methodology has also been developed for sequence labelling and sequence generation which is closer to our ideas on language decoding.~\cite{settles2008analysis} use token label entropy with query-by-committee and \cite{deng2018adversarial} use adversarial training. However, \cite{deng2018adversarial} require large number of labelled examples. Our work is also related to diverse sequence generation in NLP~\cite{tam2020cluster, ippolito2019comparison, kriz2019complexity}. Overall, the idea is to perform clustering for selecting diverse sentence candidates which is closer to our proposed acquisition function - Semantics Aware Sequential Entropy (SASE). A good comparison study for diverse decoding methods are provided in \cite{ippolito2019comparison}. 


\section{Active Learning}
\subsection{Active Learning Problem Formulation}
Supervised machine learning methods require labelled training set for learning the parameters of the model as shown in Figure~\ref{fig:msase} (a). In active learning, we focus on the trade-off between the labelling budget and model performance. Below, we formally define the problem setting. 

Consider the data generating distribution $\mathcal{P}$ and a loss function $\ell:\mathbb{R}^{|\mathcal{Y}|}\rightarrow \mathbb{R}^{|\mathcal{Y}|}$  where $|\mathcal{Y}|$ is the dimension of the output space. $(x,y)$ represent a sample from $\mathcal{P}$ where $x \in \mathbb{R}^d$ and $y \in \mathbb{R}^{|\mathcal{Y}|}$. The domain of $x$ and $y$ are represented as $\mathcal{X}$ and $\mathcal{Y}$.
Let us assume that there exists an unlabelled dataset of size $m$ sampled from the data generating distribution  $\mathcal{P}$ where labels are not accessible without paying a cost for querying every single label. We denote this unlabelled set as $U_{\rm{ul}} = \{(x_1, x_2, ..., x_m)\}$. We are also provided a budget $B \ll m$, which is the maximum number of labels which can be queried in each iteartion.  Given an index set of size $B$  denoted by $S$ which is the subset of the set $I = \{1,2,\cdots,m\}$, i.e. $S \subset I$ such that $|S| = B$, the goal of active learning is to:

\begin{align}
    &\min_{f \in \mathcal{F}} \mathbb{E}_{(x,y)} [\ell(f(x),y)] \notag \\
    \text{such that } &f = \min_{S \subset I~\rm{s.t~} |S |=B} \frac{1}{B} \sum_{j=1}^{B} \ell(f(x_{s_j}), y_{s_j}), \label{eq:active_learning}
\end{align}
where $S = \{s_1,\cdots, s_B\}$ and $\mathcal{F}$ is the hypothesis class of function of interests i.e. neural networks. It is also important to note that while solving the above optimization problem, it is strictly prohibited to query more than $B$ labels.  It is clear from the above equation that we need to get the optimal index set $S$ which is an extremely hard problem. Hence, further simplification of the problem has been proposed where a small fraction of points are allowed to be queried to get more informed querying strategy subsequently. We will denote this labelled training set  as $U_
{\rm{l}}$. 

\subsection{Acquisition Function}

Fundamental principle behind modern active learning approaches lie in designing an acquisition function which can query information-rich samples from the unlabelled set. More formally, for a given model $f \in \mathcal{F}$ and unlabelled pool $U_{\rm{ul}}$, an acquisition function $a:\mathcal{F}\times \mathbb{R}^{d} \rightarrow \mathbb{R}$ decides which samples to query next as following,
\begin{align*}
    x^\star = \arg \max_{x \in U_{\rm{ul}}} a(f,x).
\end{align*}
Uncertainty based acquisition functions have become one of the most popular approach because of it's
performance and simplicity~\cite{gal2017deep}. 
There have been multiple ways of estimating uncertainty in the prediction of a machine learning models. Entropy based uncertainty prediction has its root in information theory \cite{shannon1948mathematical}. Let us assume that a probabilistic model generates predictions in the form of probability distributions $p(\cdot|x)$ on $\mathcal{Y}$ for $x\in \mathcal{X}$ then acquisition function for  \textbf{\textit{Entropy-based method} \cite{holub2008entropy,monarch2021human,nguyen2021measure}}   can be written as: 

\begin{align*}
a(x,f) \propto \sum_{y\in \mathcal{Y}} p_{\theta} (y|x) \log p (y|x).  
\end{align*}

Other ways to estimate uncertainty in model prediction includes \textit{Margin of confidence sampling} and \textit{Least confidence sampling} \cite{monarch2021human,nguyen2021measure}. In recent works, it was also proposed to incorporate diversity of samples while querying to avoid querying the labels for similar data points~\cite{ash2019deep,hacohen2022active}. In this work, we utilize entropy based method along with diversification on the space of model prediction. We discuss this in the details in the next section. 

Given this formulation, at each iteration of Active Learning, samples from unlabelled pool are queried using the acquisition function and sent for manual annotation. The annotated samples are then added to the labelled set for training the model in supervised fashion. We show this active learning loop in Figure~\ref{fig:msase} (b).

\begin{figure*}[h]
\begin{minipage}[b]{0.24\linewidth}
  \centering
  \centerline{\includegraphics[width=\linewidth]{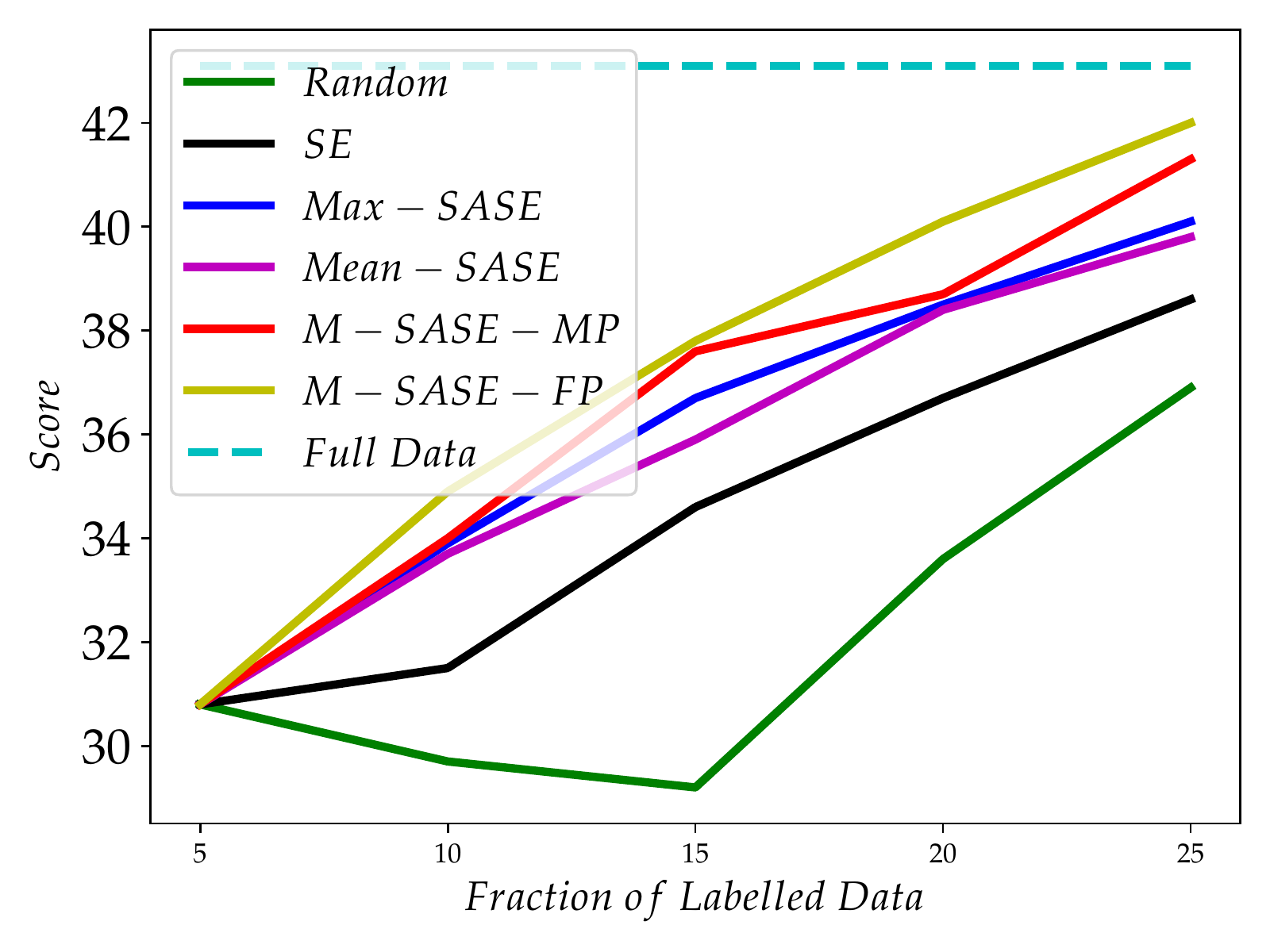}}
  \centerline{\small (a) BLEU4}
\end{minipage}
\begin{minipage}[b]{.24\linewidth}
  \centering
  \centerline{\includegraphics[width=\linewidth]{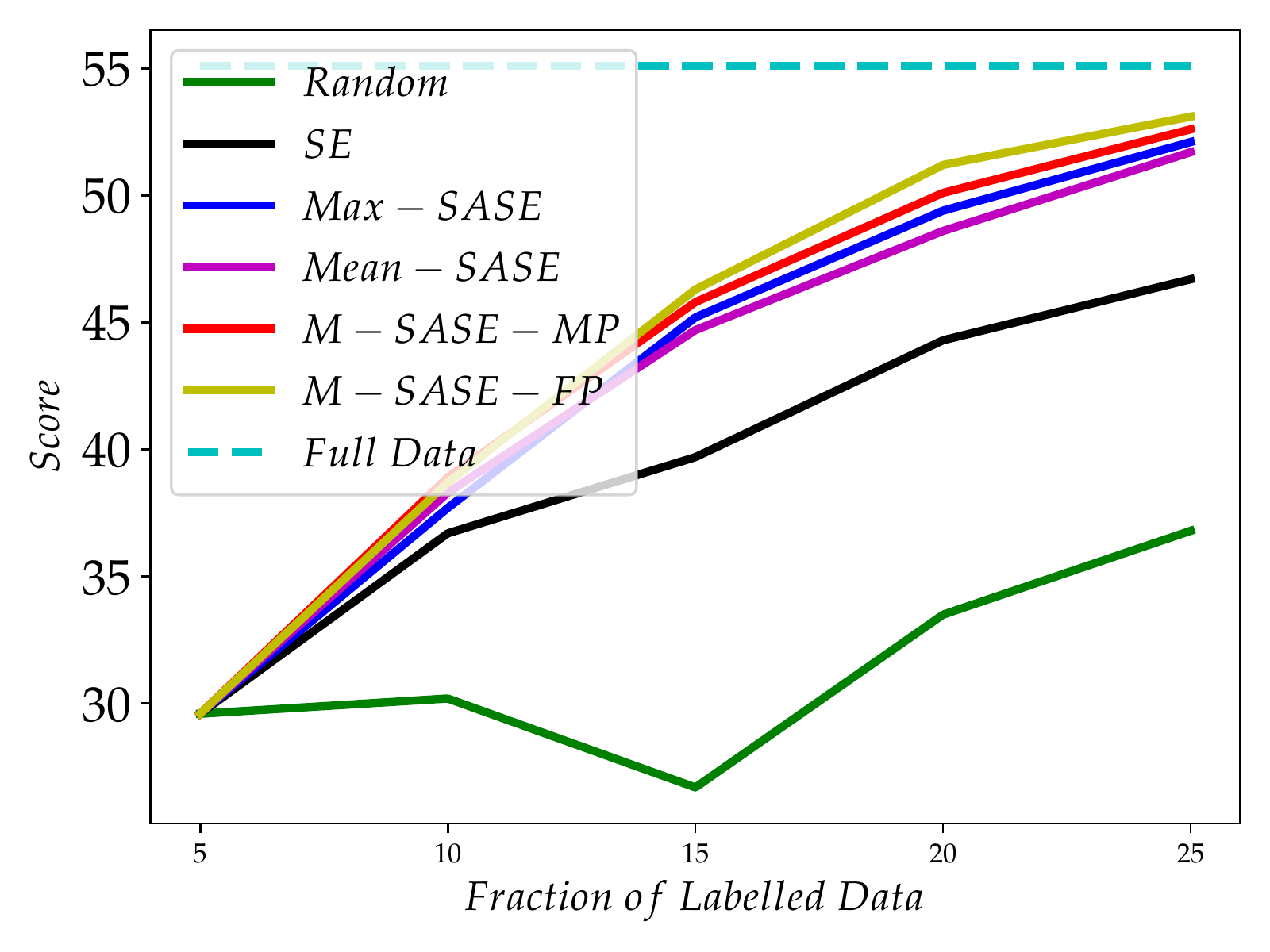}}
  \centerline{\small (b) CIDEr}
\end{minipage}
\begin{minipage}[b]{0.24\linewidth}
  \centering
  \centerline{\includegraphics[width=\linewidth]{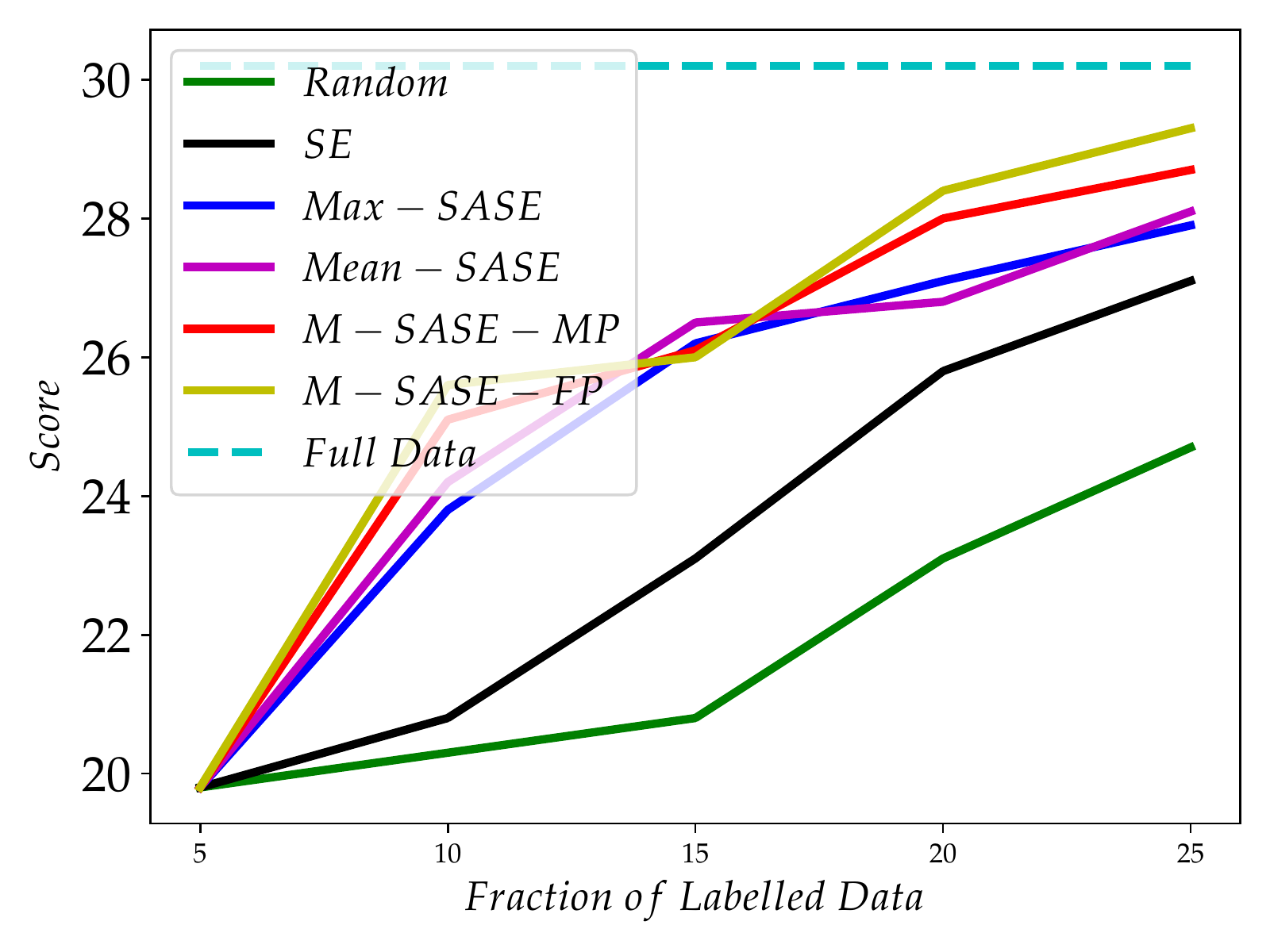}}
  \centerline{\small (c) METEOR}
\end{minipage}
\begin{minipage}[b]{0.24\linewidth}
  \centering
  \centerline{\includegraphics[width=\linewidth]{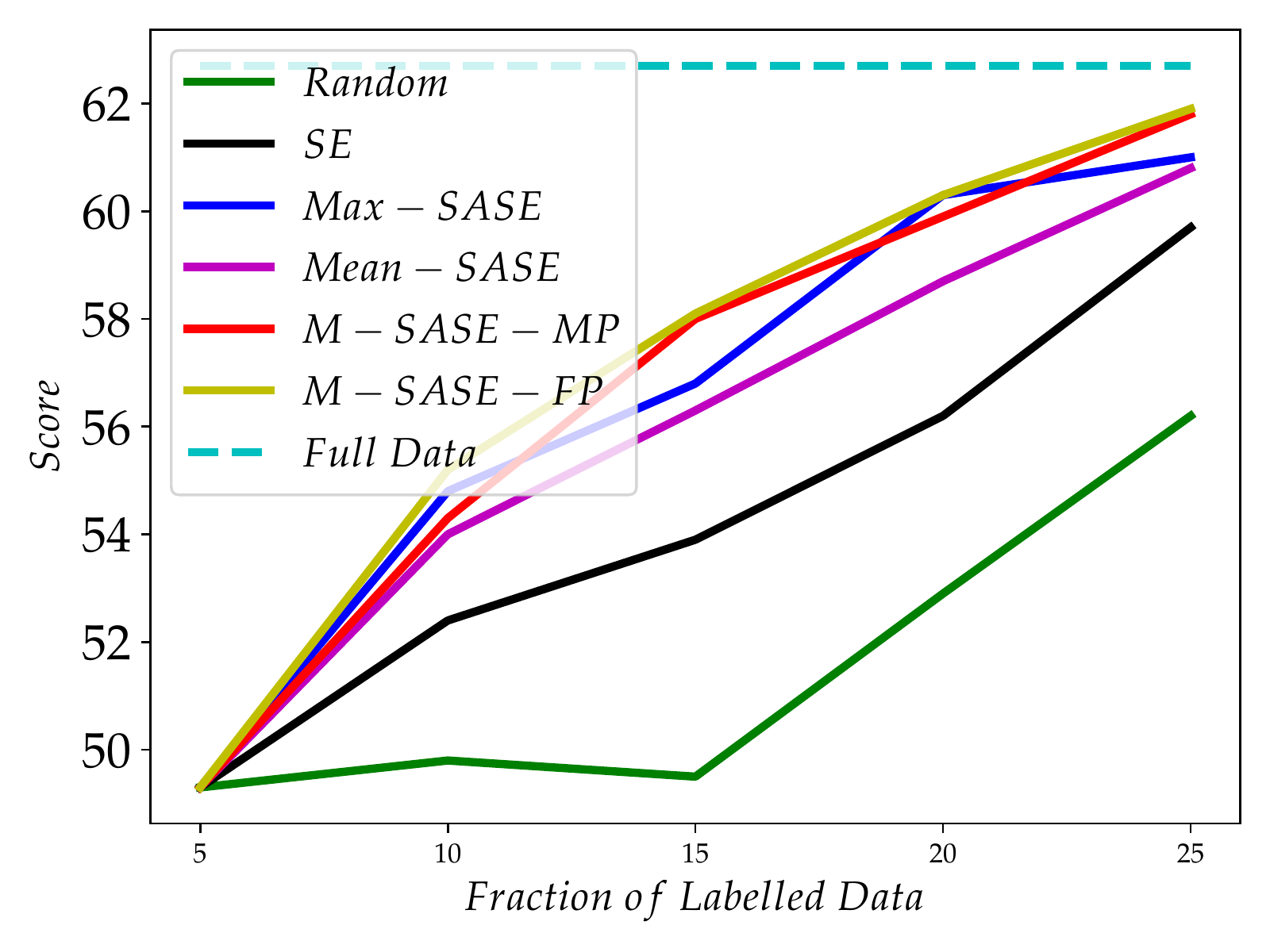}}
  \centerline{\small (d) ROUGE-L}
\end{minipage}
\vspace{-2mm}
\caption{Results of proposed approach on MSR-VTT dataset. We vary the amount of training data in steps of 5\% till 25\% of data is consumed and report the BLEU4, CIDEr, METEOR and ROUGE-L results on the test set.}
\label{fig:msrvtt_results}
\vspace{-4mm}
\end{figure*}

\section{MAViC: Multimodal Active Learning}
In this section, we discuss active learning for video captioning task followed by discussion on our approach. 

\subsection{Active Learning for Video Captioning}
In the context of active learning for video captioning, the label $y_i$ is sequential in nature as it contains textual descriptions. The parameter of video captioning system, $\theta$ comprise of two parts i)\textit{Visual Encoder} and ii) \textit{Language Decoder}, which we will denote $\theta_v$ and $\theta_l$ respectively. For every video $x$, we denote the output of the video encoder as $f_{v}(\theta_v;x)$ and final output of the language decoder is denoted as $f_l(\theta_l; f_{v}(\theta_v;x))$ which we also denote by $f(\theta;x)$. It is also important to note that $f_l$ has sequential nature and can further be decomposed as token level probability. Hence, we will denote a caption generator model with parameter $\theta = [\theta_v; \theta_{l}]$ as a triplet $(f_{\theta}, f_{v_{\theta_v}}, f_{l_{\theta_l}} )$. For video captioning task, the goal of the active learning remain similar as Equation~\eqref{eq:active_learning} where we want to train the caption generator triplet $(f_{\theta}, f_{v_{\theta_v}}, f_{l_{\theta_l}} )$ with only $B$ labelled examples. 

However, the difficulty lie in the combinatorial nature of output space. For example, let us assume that there are $D$ number of words in the dictionary and the number of tokens in each caption is fixed to $K$. Then, number of possible captions generated can be $D^K$, where  many of them would be semantically very similar and many of them are not realistically possible. Hence, designing an acquisition function for this task is not as easy as that for classification task with finite number of labels. However, we take inspirations from the design of existing acquisition functions for classification task and we utilize extra multimodal information to devise active learning method for video captioning. 
 
\subsubsection{Sequential Entropy (SE)} 
\label{sec:seq_entr}
We first discuss the entropy based active learning approach on the output of the video captioning model $f(\theta;x)$. We consider this approach as the baseline. For every video input $x$, the language decoder generates top $k$ captions selected using beam search. 
Let us denote the output for video input $x$ as $y_i$ for $i \in \{1, 2, \cdots k\}$. $y_i$ is sequential in nature and consists of words where $j$-th word for $i$-th caption is denoted as $y_{ij}$. For all top $k$ candidates, $y_i$, we compute the log-likelihood score $s_{y_i}$ and convert these scores to probabilities $p_{y_i}$ by normalizing them. We denote the score as $p_{y_i}$ for $i \in \{1,2,\cdots, k\}$. Hence, for video input $x$, the entropy of the random variable sampled from the output distribution which has discrete support on $\{y_1,\cdots,y_k\}$ can be written as, 
\begin{align*}
    H(x) = - \sum_{i=1}^k p_{y_i} \log p_{y_i}.
\end{align*}

We perform our computation on top k captions only because the output space of the caption generation algorithm is exponentially large, making it challenging to compute uncertainty over the complete output space. Similar efforts towards sequential entropy were also explored in ~\cite{chan2020active}. In the first approach,~\cite{chan2020active} averages the word-level entropy to obtain the caption-level entropy and selects the videos showing highest entropy. However, since the caption probability is the joint probability over words, average entropy over words is not able to capture the dependence and results in inferior performance. In the second approach,~\cite{chan2020active} averages over the likelihood score and selects the samples with minimum likelihood. Likelihood-based method showed improved performance over word-level entropy. However, neural networks can fail with high confidence, so the absolute value of likelihood is not a reliable measure for model confidence. Thus, in our work we compute the entropy over likelihood score to capture the gaps of the two approaches proposed by~\cite{chan2020active}. Entropy over the likelihood score normalizes the absolute value of likelihood and makes it possible to compare across videos. We also tried experiments with using mean and standard deviation over likelihood score, but obtained inferior performance as shown in Table~\ref{table:entropy_mean_sd}.

\subsubsection{Semantics Aware Sequential Entropy (SASE)} \label{sec:enhanced_seq_entr}

While sequential entropy (SE) is a natural extension of entropy to sequences, it has some limitations when applied to video captioning task. Video captioning systems have a tendency to output multiple semantically similar captions with high probabilities. For example, for an input video, consider the following two candidates - ``\textit{A person is jogging}'' with score $s_1$ and ``\textit{A man is running}'' with score $s_2$. If the score $s_1 \approx s_2$, in the current sequential entropy formulation, the output entropy of the model prediction would be high. That would imply that the entropy of the system for this example is high, leading to querying of this example. However, this high entropy could be due to multiple semantically similar and correct captions, instead of model uncertainty. Due to this complex nature of natural language captions, SE assigns high entropy to the samples even though they are not informative and model is not confused about them. As a consequence, we end up in querying the captions for videos for which the system has generated semantically similar sentences with high probability. 

To address this, we employed diversity while computing entropy to suppress high entropy due to presence of semantically similar captions. 
As shown in Figure~\ref{fig:msase} (d), for generated captions $\{y_1, \cdots, y_K\}$, we compute semantic embedding for each captions using BERT model~\cite{devlin2018bert}. Let us denote the embedding generated by the BERT model for caption $y_i$ by $b_{y_i}$. Once we have the embedding vectors $\{b_{y_1}, \cdots b_{y_k}\}$, we perform clustering on these vectors so that the captions which are semantically similar are grouped together to have one single contribution while calculating entropy. We fix the number of clusters $C$ and the index set $i_c$ denote the index set for the captions which belongs to cluster number $c$ after performing K-Means clustering. We further represent the likelihood score for the cluster $c$ in the following two ways:
\begin{itemize}
    \item \textbf{Maximum SASE: } For each cluster $c$, we represent the likelihood score as,
    \begin{align*}
        \rm{score}[c] = \max_{i \in i_c}~~ \rm{s}_i,
    \end{align*}
    where $\rm{s}_i$ denotes the likelihood score for $i$-th caption.  Basically, we select the caption with highest likelihood score from each cluster and discarding all the other captions from the cluster.
    
    \item \textbf{Mean SASE: } For cluster $c$, we represent the likelihood score as,
    \begin{align*}
        \rm{score}[c] = \frac{1}{|i_c|}\sum_{i \in i_c} \rm{s}_i,
    \end{align*}
    where $\rm{s}_i$ denotes the likelihood score for $i$-th caption. This is equivalent to having the joint likelihood of all the captions in the cluster considering independence between captions. For an input video sample $x$, as can be seen,
    \begin{align*}
        \rm{score}[c] \propto \sum_{i \in i_c} \rm{s}_i  = \sum_{i \in i_c} \log  p(y_i|x) = \log \prod_{i \in i_c} p(y_i|x).
    \end{align*}
\end{itemize}
Now, that we have likelihood score for each cluster, we can compute the probability as $p_c \propto \exp{(\rm{score}[c])} $. Hence, enhanced entropy can be computed as: $$H(x) = -\sum_{c=1}^C p_c \log p_c,$$ where $C$ is number of clusters.

    

\subsection{Multimodal SASE (M-SASE)}
Both Sequential Entropy (Section~\ref{sec:seq_entr}) and Semantics Aware Sequential Entropy (Section~\ref{sec:enhanced_seq_entr}) measure the entropy of the language decoder without considering the visual encoder. In this section, we will discuss how to utilize the information from the visual encoding in our algorithm. The key idea behind active learning algorithm lies in estimating the uncertainty in the predictions of the model. Given that video captioning consists of a visual encoder as well as language decoder, we leverage the uncertainty in the visual encoder also. 
Below, we discuss two approaches to estimate uncertainty (i) \textit{Feature perturbation (FP)} and, (ii) \textit{Model perturbation (MP)}. 

\subsubsection{Feature Perturbation (FP)}
\label{sec:msase-fp}



 \noindent\textbf{Adversarial Robustness:} We take inspiration from adversarial robustness literature \cite{silva2020opportunities,ducoffe2018adversarial} to perturb the features of the model to produce different outputs. The main idea behind adversarial robustness based active learning for classification lies in finding an optimal perturbation vector of small magnitude $\varepsilon$ which could lead to change in predictive label. Change in predictive label of the input for small $\varepsilon$, implies high uncertainty in model prediction. 

However, unlike robust classification task, which has a taxonomy of categories, output space of video captioning is open-set. While the vocabulary is predefined, the generated captions are unconstrained/open-set making it challenging to formulate the \textit{category} of a caption and thus the optimal adversarial perturbation direction which changes this \textit{category}. Hence, we utilize unsupervised clustering for this purpose. We divide the embedding space of the output of visual encoder in multiple clustered regions using K-Means clustering and consider the cluster centers as \textit{pseudo-labels}. We then perturb the visual embeddings towards $K$-nearest cluster centers and generate captions corresponding to these perturbed visual embeddings. We apply SASE on these captions to measure the information of the video example.

Formally, let us denote the labelled set $U_{\rm{l}}=\{(x_i,y_i)\}_{i=1}^n$ where $(x_i,y_i)$ is $i$-th video and its corresponding caption pair. To obtain the perturbation direction, we perform unsupervised  clustering on the output of video encoder and obtain $C$ number of clusters which is denoted as $B_1,\cdots B_C$. Let us denote the cluster centers corresponding to cluster $B_p$ by $c_p$  and the index set of points belonging to $B_p$ as $I_p$ for $p \in \{1, \cdots C\}$. For an unlabelled video input $x_i^{u}$, we perturb it towards the center of $K$- nearest cluster centers and generate $K$-output of the video encoder for input $x$ as follows,
\small
\begin{align}
    f_v^{\rm{per}}(\theta_v;x_i^{{u}})[k] = (1-\varepsilon) f_v(\theta_v;x_i^{{u}}) + \varepsilon \cdot \frac{c_{i_k} -f_v(\theta_v;x_i^{{u}}) }{\|f_v(\theta_v;x_i^{{u}})- c_{i_k}\|_2}, \label{eq:feat_perturb}
\end{align}
\normalsize
 for $k \in \{1,\cdots,K\}$ where $c_{i_k}$ represents $k$-th nearest cluster center to the $f_v(\theta_v;x_i^{{u}})$ and $\varepsilon$ is the $\ell_2$ norm of the perturbation vector. Now, that we have have $K$ perturbed output of the video encoder corresponding to the unlabelled video input $x_i^{{u}}$. We pass each perturbed input through the language decoder to generate diverse set of captions. Further, we deploy methodology discussed in sections~\ref{sec:seq_entr} and \ref{sec:enhanced_seq_entr} to estimate uncertainty via entropy over the generated captions.

\subsubsection{Model Perturbation (MP)}
\label{sec:msase-mp}
Similar to feature perturbation, we also apply model perturbation on the visual encoder. Dropout based Monte-Carlo (MC) sampling has been used extensively to measure the uncertainty of deep models~\cite{gal2016dropout, seoh2020qualitative}. The key idea is perform multiple forward passed $L$ through the model $\theta$ to generate $L \times K$ different captions without deactivating the dropout layer.  Since, each video inference selects a random value of dropout, it generates a different set of $K$ candidate captions $Y_l \in \{y_1, y_2, ... y_k\}$. Intuitively, this is similar to having an ensemble of video captioning models predicting different captions. We accumulate all these candidate captions for each video and apply SASE over these candidates and select the videos with captions showing maximum entropy. Considering, deep learning models require much computation and memory to train multiple models, dropout based ensemble provides an efficient alternative as we do not need to train multiple models.

\paragraph{{Algorithm}}
In previous two sections, we discussed methods to utilize multimodal information for estimating entropy. Now we provide formal active learning algorithm proposed in this paper below in Algorithm~\ref{alg:mavic}. 
As described earlier, we denote a caption generator with parameter $\theta=[\theta_v;\theta_l]$ as a triplet $(f_{\theta},f_{v_{\theta_v}},f_{l_{\theta_l}})$. 
\begin{algorithm} 
\caption{\textbf{MAViC}: Multimodal Active Learning for Video Captioning}\label{alg:mavic}
\begin{algorithmic}[1]
  \STATE \textbf{Input}: Labelled Training Data $U_{\rm{l}} = \{(x_i,y_i)\}_{i=1}^{n}$, budget $B$,  Unlabelled Pool $U_{\rm{ul}} = \{x_i^{u}\}_{i=1}^m$, epoch $L$,
  perturb $P$, Cluster $C$ and beam width $\kappa$.
  \STATE Train a caption generator $(f_{\theta},f_{v_{\theta_v}},f_{l_{\theta_l}})$  on  $U_{\rm{l}}$.
  \WHILE{$\ell$ $\leq \ell$}
   \FOR{Each samples $x_i^{(u)}$ in $U_{\rm{ul}}$}
    \STATE Captions $\gets$ Generate top $\kappa$ captions ($f(\theta;x)$).
    \STATE  Embed $\gets$ BERT(Captions)
    \FOR{$p \leq P$ }
      \STATE perturb[p] $\gets$ Obtain perturb visual features \label{lin:perturb}
      \STATE PCaptions $\gets$ $f_{l}(\theta_l;\rm{perturb}[p])$ (Top $\kappa$ captions)  
      \STATE PEmbed  $\gets$ BERT(PCaptions)
      \STATE Embed $\gets $ $[\text{Embed};~ \text{PEmbed}]$
    \ENDFOR
    \STATE Cluster Embed in C clusters. \label{lin:cluster}
    \STATE Assign scores to each cluster and compute probabilities $p_c$ \label{linL:scores}
    \STATE Compute Entropy $H(x^l) = -\sum_{c=1}^{C} p_c \log p_c$.
    \ENDFOR
    \STATE Rank $H(x)$ for all $x$ in $U_{\rm{ul}}$.
    \STATE Pick top $B/L$ samples and denote it set with $S$.
    \STATE Query label for all samples in $S$.
    \STATE  $U_{\rm{ul}}\gets U_{\rm{ul}} - S$ and $U_{\rm{l}} \gets U_{\rm{l}} \cup S $.
    \STATE Update $\theta$.
  \ENDWHILE
  \STATE \textbf{return} $\theta$.
\end{algorithmic}
\end{algorithm}
In line~\ref{lin:perturb} of algorithm~\ref{alg:mavic}, $\rm{perturb}[j]$ is obtained using both the methods discussed above (i) feature perturbation \eqref{eq:feat_perturb} and (ii) model perturbation. We use $K$-means clustering to cluster the caption embedding (Line~\ref{lin:cluster} of Algorithm~\ref{alg:mavic}). And in line~\ref{linL:scores} of Algorithm~\ref{alg:mavic}, we assign scores to each cluster using max-score and average score as discussed in section~\ref{sec:enhanced_seq_entr}.

\begin{figure*}[h]
\begin{minipage}[b]{0.24\linewidth}
  \centering
  \centerline{\includegraphics[width=\linewidth]{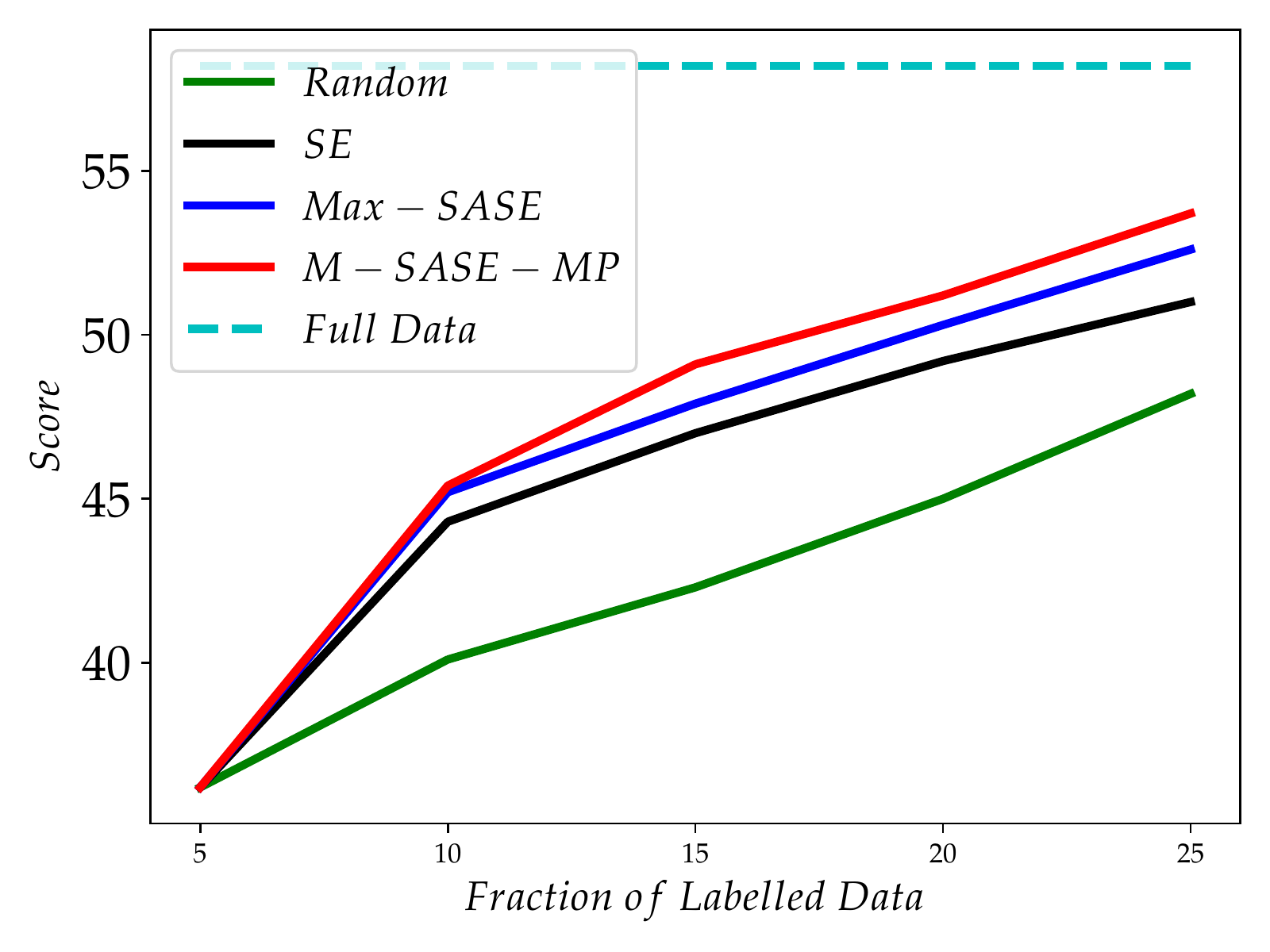}}
  \centerline{\small (a) BLEU4}
\end{minipage}
\begin{minipage}[b]{.24\linewidth}
  \centering
  \centerline{\includegraphics[width=\linewidth]{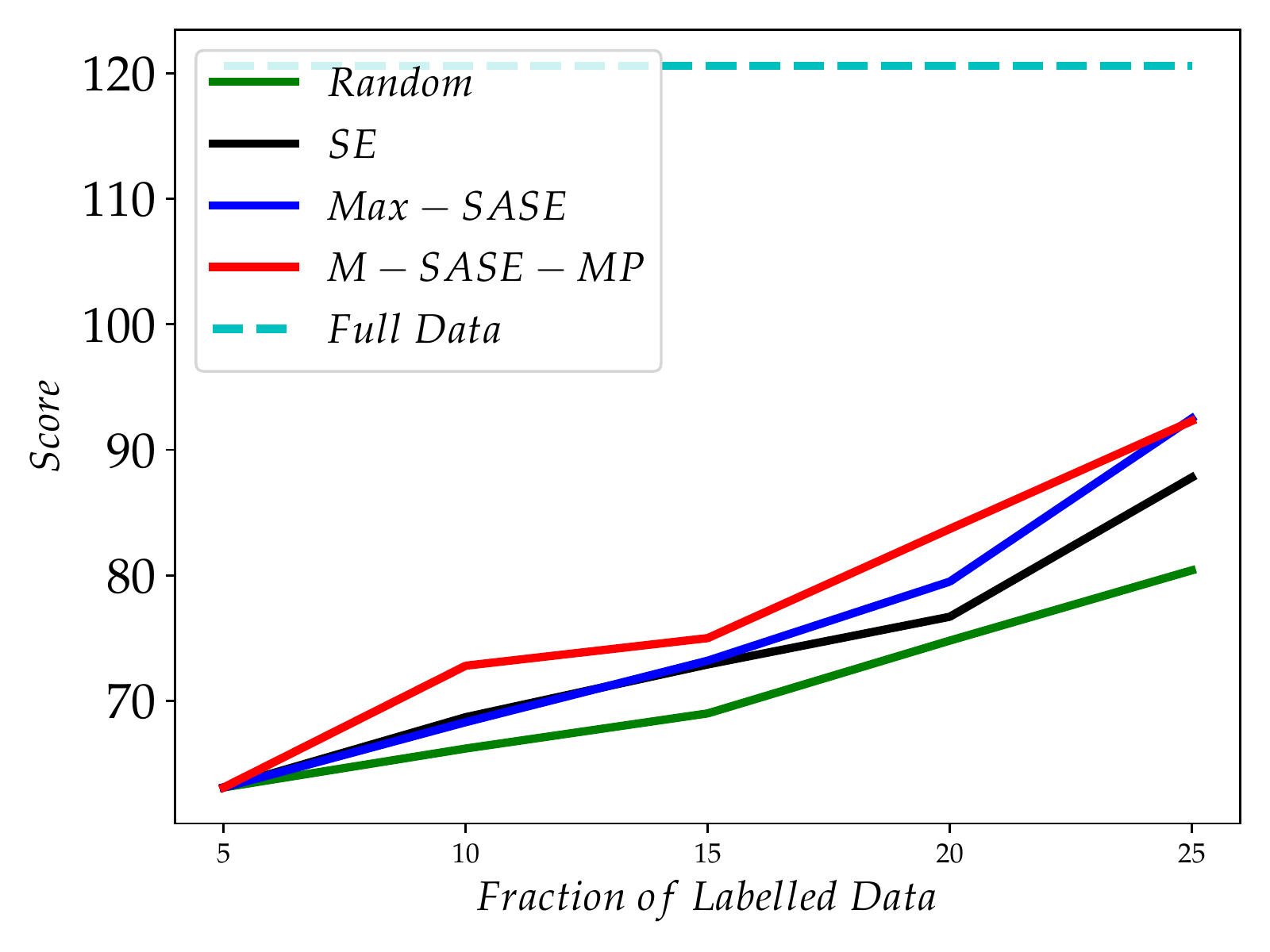}}
  \centerline{\small (b) CIDEr}
\end{minipage}
\begin{minipage}[b]{0.24\linewidth}
  \centering
  \centerline{\includegraphics[width=\linewidth]{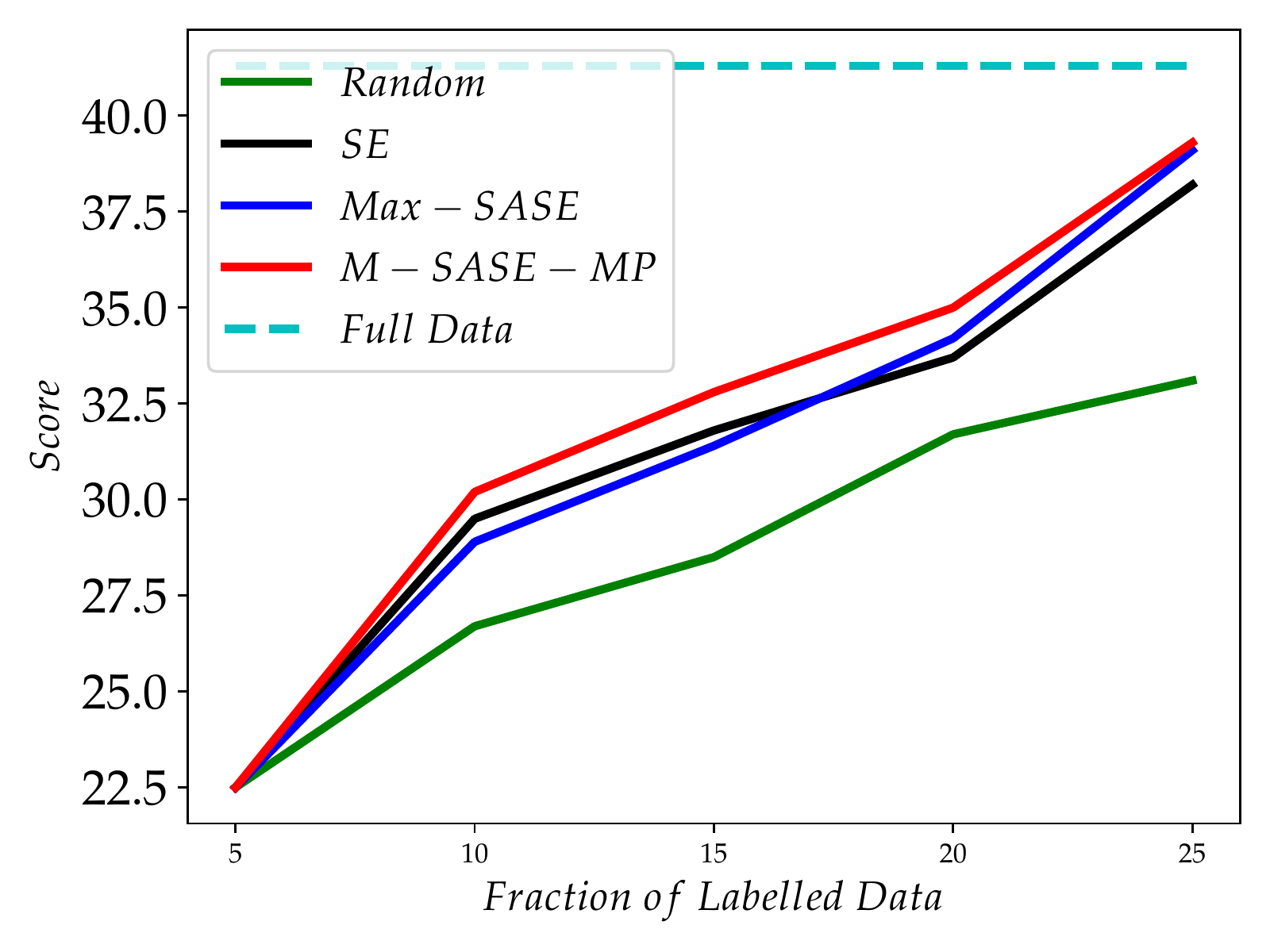}}
  \centerline{\small (c) METEOR}
\end{minipage}
\begin{minipage}[b]{0.24\linewidth}
  \centering
  \centerline{\includegraphics[width=\linewidth]{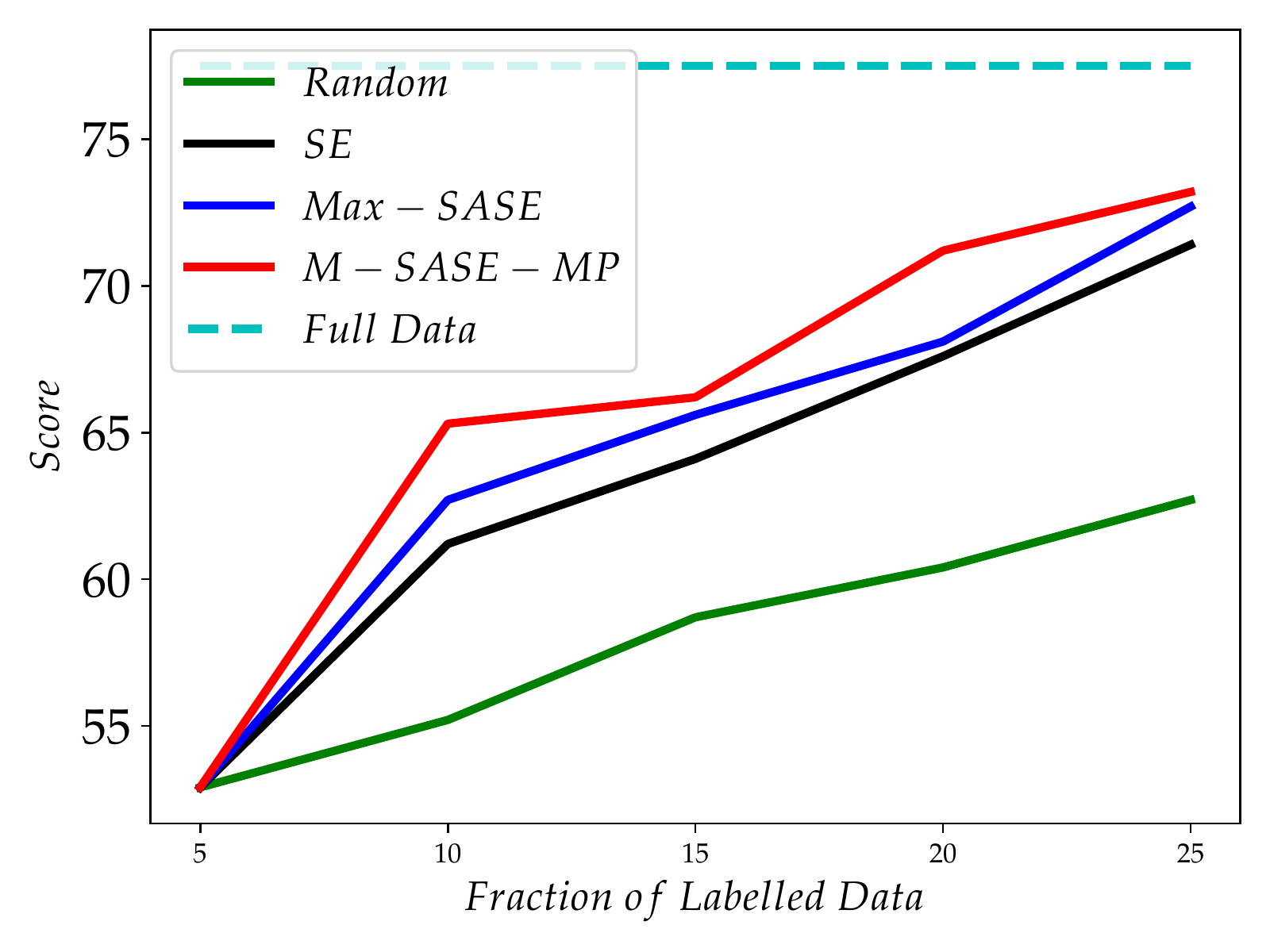}}
  \centerline{\small (d) ROUGE-L}
\end{minipage}
\vspace{-2mm}
\caption{Results of proposed approach on MSVD dataset. We vary the amount of training data in steps of 5\% till 25\% of data is consumed and report the BLEU4, CIDEr, METEOR and ROUGE-L results on the test set.}
\label{fig:msvd_results}
\vspace{-4mm}
\end{figure*}

\section{Experiments}
\subsection{Methodolgy}
In Active learning, it is ideal to sample one data point at each active learning iteration and train the complete model before requesting the next sample. However, due to large training and inference time of deep models, this is not feasible. Thus, following~\cite{chan2020active}, we randomly select 5\% of data from the labelled set and use this as the seed dataset for training the base model. We use same set of seed examples across all experiments for a dataset for fair comparison. After training the model with the base model for 15 epochs, we apply the different acquisition functions to select the next set of 5\% examples and add them to the previous set. We then finetune the model with this accumulated dataset for same number of epochs. We iterate these steps till 25\% of the dataset is consumed. At each step, we compute the performance of the model on the test set and report the results.

\subsection{Models and Datasets}
\begin{itemize}
    
\item \textbf{Model:} We used SwinBERT~\cite{lin2022swinbert} for video captioning as the baseline model. SwinBERT is state-of-the-art end-to-end transformer based video captioning model. SwinBERT consists of VidSwin~\cite{liu2022video} (pretrained with Kinetics-600) encoder and BERT-based transformer language decoder. In the decoding phase, we use number of beams and beam depth as 10 for all our experiments. We employ AdamW optimizer and use a learning rate of 1e-4 with warm-up during the early 10\% training steps followed by linear decay with batch size of 2 (MSR-VTT) and 6 (MSVD) on 4 T4 GPU for 15 epochs. We resize the shorter side of all the video frames to 224. During training, we random crop (224×224) at the same location for all the frames in a given video. During inference, we center crop (224 × 224) all the frames.

We evaluate our methods on the MSR-VTT~\cite{xu2016msr} dataset and MSVD~\cite{chen2011collecting} dataset.

\item \textbf{MSR-VTT}: MSR-VTT dataset is an open-domain dataset and consists of more than 7K videos and 10K clips of approximately 20 seconds duration sourced from YouTube. The dataset contains 200K clip-caption pairs with each clip being annotated with 20 captions using Amazon Mechanical Turk (AMT). We use the standard splits and consider 6.5K clips as training set and 2.9K clips as the test set. 

\item \textbf{MSVD}: MSVD is also sourced from YouTube and contains 1970 clips with average duration of around 10 seconds. Each clip has been annotated with one caption using Amazon Mechanical Turks (AMT). We use the standard splits with 1200 train, 100 validation and 670 test videos.

\item \textbf{Evaluation Metrics:} We measure the performance of our models using - BLEU4~\cite{papineni2002bleu}, METEOR~\cite{banerjee2005meteor}, ROUGE-L~\cite{lin2004automatic} and CIDEr~\cite{vedantamconsensus}.

\end{itemize}

\subsection{MAViC Results}
\begin{itemize}

\item \textbf{Random Sampling:} We consider randomly sampling 5\% samples at each step as our baseline approach and report the results in Figure~\ref{fig:msrvtt_results} and Figure~\ref{fig:msvd_results} for MSR-VTT and MSVD datasets. Random sampling mirrors the distribution of the dataset without incorporating the signals from the trained model. As expected, the performance of the model improves on adding more samples for training the models.  

\item \textbf{SE Experiments:} We replace the random sampling with Sequential Entropy (SE) in these experiments. At each iteration, we select samples from the unlabelled dataset which exhibit higher sequential entropy using the model trained till this iteration. We observe substantial gains over \textit{random sampling} baseline which shows the efficacy of using sequential entropy over random sampling as shown in Figure~\ref{fig:msrvtt_results} and Figure~\ref{fig:msvd_results}. 

\item \textbf{SASE Experiments:} In SASE, we further enhance the sequential entropy with semantic information and observe substantial gains as shown in Figure~\ref{fig:msrvtt_results} and Figure~\ref{fig:msvd_results}. Using validation, we found the optimal number of clusters as 10 and used them in this experiment. These results show the importance of integrating semantic information in the acquisition function. Even with 15\% of data, we are able to reach the performance of SE with 10\% of data.

\item \textbf{Multimodal SASE:}
In Figure~\ref{fig:msrvtt_results} and ~\ref{fig:msvd_results}, we report the performance by using model and feature based perturbation on the visual encoder. We observe at all the data ratios, Multimodal SASE outperforms all the other results. Since, M-SASE-MP has multiple forward passes which increases the number of candidates, we increase the number of clusters to 30. Figure~\ref{fig:msrvtt_results} and Figure~\ref{fig:msvd_results} present the overall pattern. We report the detailed results in Appendix Section~\ref{section:msr_vtt_detailed} and Appendix Section~\ref{section:msvd_detailed}.

\end{itemize}

\begin{table}[h]
  \centering
  \small
  \begin{tabular}{lcccc}
    \toprule
    Method & BLEU4 & METEOR & ROUGE-L & CIDEr \\
    \midrule
    Mean & 31.2 & 20.1 & 51.0 & 33.6 \\
    Entropy & \textbf{31.5} & \textbf{20.8} & \textbf{52.4} & \textbf{36.7}\\
    \bottomrule
  \end{tabular}
  \caption{Comparison between entropy, mean and standard deviation for SE on MSR-VTT dataset}
  \label{table:entropy_mean_sd}
\end{table}

\begin{table}[h]
  \centering
  \begin{tabular}{cccccccc}
    \toprule
    Clusters & 10 & 15 & 20 & 25 & 30 & 35 & 40 \\
    \midrule
    CIDEr & 36.3 & 37.1 & 37.6 & 38.5 & \textbf{38.9} & 38.3 & 35.6\\
    \bottomrule
  \end{tabular}
  \caption{Model performance (CIDEr score) on varying number of clusters for M-SASE on MSR-VTT dataset}
  \label{table:ablation_num_k}
\end{table}

\begin{table}[h]
  \centering
  \begin{tabular}{ccccc}
    \toprule
    Method & BLEU4 & METEOR & ROUGE-L & CIDEr \\
    \midrule
    w/o FT & 31.2 & 20.5 & 51.7 & 34.0 \\
    FT & \textbf{31.5} & \textbf{20.8} & \textbf{52.4} & \textbf{36.7} \\
    \bottomrule
  \end{tabular}
  \caption{Comparison (CIDEr score) between finetuning and training from scratch (w/o finetuning) for SASE on MSR-VTT dataset on 10\% of data.}
  \label{table:ft_vs_scratch}
\end{table}

\begin{table}[h]
  \centering
  \begin{tabular}{cccccccc}
    \toprule
    Clusters & 200 & 300 & 400 & 500\\
    \midrule
    CIDEr & 37.1 & \textbf{37.5} & 36.7 & 35.3 \\
    \bottomrule
  \end{tabular}
  \caption{Model performance (CIDEr score) on varying number of clusters for M-SASE-FP on MSR-VTT dataset}
  \label{table:FP_Varying_K}
\end{table}

\begin{table}[h]
  \centering
  \begin{tabular}{cccccccc}
    \toprule
    epsilon & 0.0001 & 0.001 & 0.01 & 0.015 & 0.02\\
    \midrule
    CIDEr & 34.6 & 37.5 & 38.1 & 38.7 & 38.3\\
    \bottomrule
  \end{tabular}
  \caption{Model performance (CIDEr score) on varying epsilon for M-SASE-FP on MSR-VTT dataset}
  \label{table:FP_Varying_eps}
\end{table}

\subsection{Ablation Studies}

\begin{itemize}
    
\item \textbf{Entropy vs Mean:} We compared the efficacy of entropy over the average likelihood score in Table~\ref{table:entropy_mean_sd} for Sequential Entropy (SE)(Section~\ref{sec:seq_entr}). In average likelihood experiments, we compute likelihood of unlabelled samples and select the samples showing least likelihood. The intuition is that low likelihood score implies lack of confidence of model on the sample. We note that entropy shows gains over mean, so we use entropy over likelihood score for all our experiments.

\item \textbf{Forward-pass and Clusters:} For multimodal M-SASE-MP (Section~\ref{sec:msase-mp}), we perform ablation experiments to find optimal number of clusters and number of forward pass. We experiment with one iteration of active learning (5\% as seed data and 5\% acquired by M-SASE) due to compute constraints. We keep the number of forward passes fixed to 5 and vary the number of clusters from 10 to 40. From Table~\ref{table:ablation_num_k}, we observe that performance improves on increasing the number of clusters to 30 and then decreases. Fixing the number of clusters to 30, we further experimented with more number of forward passes but did not observe gains. CIDEr score dropped to 37.6 from 38.9 on increasing number of forward passes to 10. Thus, we use 5 and 30 as the number of forward passes and number of clusters respectively for the experiments.

\item \textbf{Training Strategy:} We compare the impact of finetuning the model from previous checkpoint at every iteration of active learning and training from scratch in Table~\ref{table:ft_vs_scratch}. We perform this study for 10\% of data on MSR-VTT. We note that finetuning demonstrates superior performance. Thus, for all our experiments, we continue with finetuning strategy.

\item \textbf{Feature Perturbation:} In Table~\ref{table:FP_Varying_K}, we vary the number of clusters used during clustering of visual features in M-SASE-FP (Section~\ref{sec:msase-fp}) and observe that 300 clusters give best CIDEr score. We also experiment with epsilon (Equation ~\ref{eq:feat_perturb}) in Table~\ref{table:FP_Varying_eps} and observe 0.015 as the optimal value. 


\end{itemize}

\section{Conclusion}
Developing state-of-the-art video captioning models require large number of labelled video-caption pair, which is expensive and time-consuming to collect. To tackle this, we explored active learning for video captioning in this work. We introduced a novel method \textbf{MAViC}, which utilises our proposed Semantically Aware Sequential Entropy (SASE) acquisition function to discourage querying less-informative samples which exhibit high entropy due to semantically similar captions. We also extended our approach to capture the model uncertainty in the visual dimension by feature perturbation (M-SASE-FP) and model perturbation (M-SASE-MP) and propose multimodal extension of SASE termed as M-SASE in our study. Overall, our empirical experiments show that our proposed method shows substantial improvements over the baseline approaches.

\clearpage


{\small
\bibliographystyle{ieee_fullname}
\bibliography{egbib}
}
\clearpage

\clearpage

\section{Appendix}

\section{Detailed Results - MSR-VTT}
\label{section:msr_vtt_detailed}
In this section, we report the detailed results for our experiments on MSR-VTT. These results are also plotted in main paper Figure ~\ref{fig:msrvtt_results}. In Table~\ref{table:random}, we present the results using random sampling as the baseline. In Table~\ref{table:seq_entropy}, we report the results for Sequential Entropy (SE) (Section~\ref{sec:seq_entr}). Table~\ref{table:diversity_mean_msrvtt} and Table~\ref{table:diversity_max} showcase the results for Semantically Aware Sequential Entropy (SASE) (Section~\ref{sec:enhanced_seq_entr}). Multimodal-SASE (M-SASE) results are present in the Table~\ref{table:multimodal_dropout} (M-SASE-MP) (Section~\ref{sec:msase-mp}) and Table~\ref{table:perturbation} (M-SASE-FP) (Section~\ref{sec:msase-fp}).

\begin{table}[h]
  \centering
  \begin{tabular}{ccccc}
    \toprule
    Data Ratio & BLEU4 & METEOR & ROUGE-L & CIDEr \\
    \midrule
    5\%  & 30.8 & 19.8 & 49.3 & 29.6\\
    10\% & 29.7 & 20.3 & 49.8 & 30.2\\
    15\% & 29.2 & 20.8 & 49.5 & 26.7\\
    20\% & 33.6 & 23.1 & 52.9 & 33.5\\
    25\% & 36.9 &  24.7& 56.2 & 36.8\\
    \midrule
    Supervised & 43.1 &  30.2 & 62.7 & 55.1\\
    \bottomrule
  \end{tabular}
  \caption{Results using \textit{random sampling} on MSR-VTT dataset for different data ratios}
  \label{table:random}
\end{table}


\begin{table}[h]
  \centering
  \begin{tabular}{ccccc}
    \toprule
    Data Ratio & BLEU4 & METEOR & ROUGE-L & CIDEr \\
    \midrule
    5\%  & 30.8 & 19.8 & 49.3 & 29.6\\
    10\% & 31.8 &  20.7 &  52.6 & 37.4\\
    15\% & 34.1  & 22.9 & 54.1 & 39.1\\
    20\% &  36.0 &  25.3 & 55.9 & 43.8\\
    25\% & 38.2 & 26.7 &  58.8 & 46.2\\
    \midrule
    Supervised & 43.1 &  30.2 & 62.7 & 55.1\\
    \bottomrule
  \end{tabular}
  \caption{Results using Sequential Entropy (SE) on MSR-VTT dataset for different data ratios}
  \label{table:seq_entropy}
\end{table}

\begin{table}[h]
  \centering
  \begin{tabular}{ccccc}
    \toprule
    Data Ratio & BLEU4 & METEOR & ROUGE-L & CIDEr \\
    \midrule
    5\%  & 30.8 & 19.8 & 49.3 & 29.6\\
    10\% & 33.7 & 24.2 & 54.0 & 38.3\\
    15\% & 35.9 & 26.5 & 56.3 & 44.7\\
    20\% & 38.4 & 26.8 & 58.7 & 48.6\\
    25\% & 39.8 & 28.1 & 60.8 & 51.7\\
    \midrule
    Supervised & 43.1 &  30.2 & 62.7 & 55.1\\
    \bottomrule
  \end{tabular}
  \caption{Results using Mean-SASE on MSR-VTT dataset for different data ratios}
  \label{table:diversity_mean_msrvtt}
\end{table}

\begin{table}[h]
  \centering
  \begin{tabular}{ccccc}
    \toprule
    Data Ratio & BLEU4 & METEOR & ROUGE-L & CIDEr \\
    \midrule
    5\%  & 30.8 & 19.8 & 49.3 & 29.6\\
    10\% & 33.9 & 23.8 & 54.8 & 37.7\\
    15\% & 36.7 & 26.2 & 56.8 & 45.2\\
    20\% & 38.5 & 27.1 & 60.3 & 49.4\\
    25\% & 40.1 & 27.9 & 61.0 & 52.1\\
    \midrule
    Supervised & 43.1 &  30.2 & 62.7 & 55.1\\
    \bottomrule
  \end{tabular}
  \caption{Results using Max-SASE on MSR-VTT dataset for different data ratios}
  \label{table:diversity_max}
\end{table}

\begin{table}[h]
  \centering
  \begin{tabular}{ccccc}
    \toprule
    Data Ratio & BLEU4 & METEOR & ROUGE-L & CIDEr \\
    \midrule
    5\%  & 30.8 & 19.8 & 49.3 & 29.6\\
    10\% & 34.0 & 25.1 & 54.3 & 38.9\\
    15\% & 37.6 & 26.1 & 58.0 & 45.8\\
    20\% & 38.7 & 28.0 & 59.9 & 50.1\\
    25\% & 41.3 & 28.7 & 61.8 & 52.6\\
    \midrule
    Supervised & 43.1 &  30.2 & 62.7 & 55.1\\
    \bottomrule
  \end{tabular}
    \caption{Results using M-SASE-MP on MSR-VTT dataset for different data ratios}
    \label{table:multimodal_dropout}
\end{table}

\begin{table}[h]
  \centering
  \begin{tabular}{ccccc}
    \toprule
    Data Ratio & BLEU4 & METEOR & ROUGE-L & CIDEr \\
    \midrule
    5\%  & 30.8 & 19.8 & 49.3 & 29.6\\
    10\% & 34.9 & 25.6 & 55.2 & 38.7\\
    15\% & 37.8 & 26.0 & 58.1 & 46.3\\
    20\% & 40.1 & 28.4 & 60.3 & 51.2\\
    25\% & 42.0 & 29.3 & 61.9 & 53.1\\
    \midrule
    Supervised & 43.1 &  30.2 & 62.7 & 55.1\\
    \bottomrule
  \end{tabular}
    \caption{Results using M-SASE-FP on MSR-VTT dataset for different data ratios}
    \label{table:perturbation}
\end{table}

\section{Detailed Results - MSVD}
\label{section:msvd_detailed}
In this section, we report the detailed results for our experiments on MSVD. These results are also plotted in main paper Figure ~\ref{fig:msvd_results}.

\begin{table}[H]
  \centering
  \begin{tabular}{ccccc}
    \toprule
    Data Ratio & BLEU4 & METEOR & ROUGE-L & CIDEr \\
    \midrule
    5\%  & 36.2 & 22.5 & 52.9 & 63.1\\
    10\% & 40.1 & 26.7 & 55.2 & 66.2\\
    15\% & 42.3 & 28.5 & 58.7 & 69.0\\
    20\% & 45.0 & 31.7 & 60.4 & 74.8\\
    25\% & 48.2 & 33.1 & 62.7 & 80.4\\
    \midrule
    Supervised & 58.2 & 41.3 & 77.5 & 120.6\\
    \bottomrule
  \end{tabular}
  \caption{Results using \textit{random sampling} on MSVD dataset for different data ratios}
  \label{table:random_msvd}
\end{table}

\begin{table}[H]
  \centering
  \begin{tabular}{ccccc}
    \toprule
    Data Ratio & BLEU4 & METEOR & ROUGE-L & CIDEr \\
    \midrule
    5\%  & 36.2 & 22.5 & 52.9 & 63.1\\
    10\% & 44.3 &  29.5 &  61.2 & 68.7\\
    15\% & 47.0 & 31.8 & 64.1 & 72.9\\
    20\% & 49.2 & 33.7 & 67.6 & 76.7\\
    25\% & 51.0 & 38.2 & 71.4 & 87.8\\
    \midrule
    Supervised & 58.2 & 41.3 & 77.5 & 120.6\\
    \bottomrule
  \end{tabular}
  \caption{Results using Sequential Entropy (SE) on MSVD dataset for different data ratios}
  \label{table:seq_entropy_msvd}
\end{table}

\begin{table}[H]
  \centering
  \begin{tabular}{ccccc}
    \toprule
    Data Ratio & BLEU4 & METEOR & ROUGE-L & CIDEr \\
    \midrule
    5\%  & 36.2 & 22.5 & 52.9 & 63.1\\
     10\% & 43.1 & 28.2 & 59.3 & 67.5\\
    15\% & 47.0 & 30.7 & 64.7 & 73.5\\
    20\% & 49.2 & 32.7 & 66.9 & 78.8\\
    25\% & 51.0 & 37.9 & 70.8 & 91.4\\
    \midrule
    Supervised & 58.2 & 41.3 & 77.5 & 120.6\\
    \bottomrule
  \end{tabular}
  \label{table:diversity_mean}
    \caption{Results using Mean-SASE on MSVD dataset for different data ratios}
\end{table}

\begin{table}[H]
  \centering
  \begin{tabular}{ccccc}
    \toprule
    Data Ratio & BLEU4 & METEOR & ROUGE-L & CIDEr \\
    \midrule
    5\%  & 36.2 & 22.5 & 52.9 & 63.1\\
    10\% & 45.2 & 28.9 & 62.7 & 68.3\\
    15\% & 47.9 & 31.4 & 65.6 & 73.2\\
    20\% & 50.3 & 34.2 & 68.1 & 79.5\\
    25\% & 52.6 & 39.1 & 72.7 & 92.5\\
    \midrule
    Supervised & 58.2 & 41.3 & 77.5 & 120.6\\
    \bottomrule
  \end{tabular}
  \label{table:diversity_max_MSVD}
   \caption{Results using Max-SASE on MSVD dataset for different data ratios}
\end{table}

\begin{table}[H]
  \centering
  \begin{tabular}{ccccc}
    \toprule
    Data Ratio & BLEU4 & METEOR & ROUGE-L & CIDEr \\
    \midrule
    5\%  & 36.2 & 22.5 & 52.9 & 63.1\\
    10\% & 45.4 & 30.2 & 65.3 & 72.8\\
    15\% & 49.1 & 32.8 & 66.2 & 75.0\\
    20\% & 51.2 & 35.0 & 71.2 & 83.7\\
    25\% & 53.7 & 39.3 & 73.2 & 92.3\\
    \midrule
    Supervised & 58.2 & 41.3 & 77.5 & 120.6\\
    \bottomrule
  \end{tabular}
    \caption{Results using M-SASE-MP on MSVD dataset for different data ratios}
    \label{table:multimodal_dropout_MSVD}
\end{table}

\begin{table}[H]
  \centering
  \begin{tabular}{ccccc}
    \toprule
    Data Ratio & BLEU4 & METEOR & ROUGE-L & CIDEr \\
    \midrule
    5\%  & 36.2 & 22.5 & 52.9 & 63.1\\
    10\% & 46.7 & 31.6 & 67.4 & 74.5\\
    15\% & 51.0 & 33.9 & 68.1 & 77.8\\
    20\% & 53.3 & 36.7 & 73.5 & 85.6\\
    25\% & 55.0 & 41.2 & 74.8 & 95.4\\
    \midrule
    Supervised & 58.2 & 41.3 & 77.5 & 120.6\\
    \bottomrule
  \end{tabular}
    \caption{Results using M-SASE-FP on MSVD dataset for different data ratios}
    \label{table:perturbation}
\end{table}

\label{sec:training_details}

\end{document}